# Delta6: A Low-Cost, 6-DOF Force-Sensing Flexible End-Effector

Yue Feng, Weicheng Huang, Chen Qiu, Huixu Dong, and I-Ming Chen, *Fellow, IEEE*

***Abstract* — This paper presents Delta6, a low-cost, six-degree-of-freedom (6-DOF) force/torque end-effector that combines antagonistic springs with magnetic encoders to deliver accurate wrench sensing while remaining as simple to assemble as flat-pack furniture. A fully 3D-printed prototype, assembled entirely from off-the-shelf parts, withstands peak forces above ±14.4 N and torques of ±0.33 N·m per axis; these limits can be further extended by leveraging the proposed parametric analytical model. Without calibration, Delta6 attains a 99th-percentile error of 7% full scale (FS). With lightweight sequence models, the error is reduced to 3.8% FS by the best-performing network. Benchmarks on multiple computing platforms confirm that the device's bandwidth is adjustable, enabling balanced trade-offs among update rate, accuracy, and cost, while durability, thermal drift, and zero-calibration tests confirm its robustness. With Delta6 mounted on a robot arm governed by a force–impedance controller, the system successfully performs two contact-rich tasks: buffing curved surfaces and tight assemblies. Experiments validate the design, showing that Delta 6 is a robust, low-cost alternative to existing 6-DOF force sensing solutions. Open-source site: https://wings-robotics.github.io/delta6.**

***Index Terms* — Robotic end-effector, 6-DOF force sensing, flexible parallel mechanism, antagonistic spring, cost-effective design**

## I. INTRODUCTION

ROBOTIC manipulators are increasingly engaged in human-centered tasks, ranging from clinical ultrasound scanning [1], massage therapy [2], and nasal swab collection [3] to collaborative industrial assembly [4] and household assistance [5]. In these scenarios, end-effectors must combine real-time, multi-DOF force sensing with compliance for safe, effective interaction: appropriate contact forces protect patients, compliant tooling absorbs assembly tolerances, and precise force feedback enables dexterous household tasks such as cooking [6-7]. Nonetheless, prevailing force-sensing solutions remain costly, provide limited compliance, and either demand prohibitively expensive manufacturing investment or fail to achieve the required range and accuracy, thereby constraining their broader deployment.

Yue Feng and I-Ming Chen are with Robotics Research Center (RRC), Nanyang Technological University, Singapore, 639798. (e-mail: yue011@e.ntu.edu.sg; michen@ieee.org)

Weicheng Huang is with WinGs Robotics LLC, NY, 10314 USA. (e-mail: info@wingsrobotics.com)

Huixu Dong is with the Robot Perception and Grasp Laboratory, Mechanical Engineering Department, Zhejiang University, Hangzhou 310027, China. (e-mail: huixudong@zju.edu.cn).

Chen Qiu is with Maider Medical Industry Equipment Company, Ltd., Taizhou 317607, China. (e-mail: qiuchenchallenge@gmail.com).

Metal strain-gauge sensors set the standard for 6-DOF force/torque (F/T) measurement, offering high accuracy, large load capacity, and compact size, which suit robotic grinding and machining [8]. Yet their intricate manufacturing and assembly raise costs [9], restricting use in budget-constrained settings. Their rigidity also poses safety risks in dynamic or unstructured environments; sudden force spikes in space teleoperation, for example, can trigger collisions [10]. Alternatively, MEMS capacitive [11] and piezoresistive [12] sensors also offer high-performance 6-DOF sensing. Although mass production could push their unit cost below that of strain-gauge sensors, advanced microfabrication still raises manufacturing costs and limits widespread adoption.

Several research groups have developed precise, compact multi-DOF F/T sensor alternatives for robotic applications [13–17]. Hendrich et al. [13] demonstrated a 6-DOF sensor based on optical proximity and photo-interrupter components for under US$20, but its accuracy remains modest, with worst-case errors nearing 10% of full scale; although open source, achieving the reported performance demands careful tuning during assembly, which hampers replication by users unfamiliar with the method. Similarly, Palli et al. [14] proposed a low-cost compact 6-DOF F/T sensor built from three small PCBs, each carrying a single IR LED and four phototransistors. The design achieves about 10% full-scale errors after careful calibration. Furthermore, Xiong et al. [15] embed fiber Bragg gratings in a compliant structure to sense deformation and recover forces, reaching 2.3 % full-scale error. Yet the approach is not cost-effective, as the necessary FBG interrogator alone is priced in the several-thousand-dollar range. By contrast, Al-Mai et al. [16] proposed a low-cost fiber-optic sensor whose bill of materials is roughly US$30 while still achieving a wide measurement range and high accuracy. However, the design requires each 0.75 mm PMMA fiber to be precisely aligned to a fixed 3.5 mm optical gap during assembly; any misalignment, long-term drift, or material ageing can degrade performance and therefore necessitate periodic recalibration. In addition, Ouyang and Howe [17] present a lower-cost vision-based approach: fiducial markers affixed to a compliant structure are tracked by a single RGB camera to estimate six-axis forces, and the hardware bill of materials, excluding the host computer, totals about US$50. The camera's frame-rate limitation, however, restricts the sensor bandwidth to 25 Hz. Other often discussed low-cost options, such as conductive-fabric sensors, suit wearables rather than robot end-effectors because of their limited range, drift, and mostly single-axis response [18].

In this context, cost-effective, easy-to-build-and-maintain,

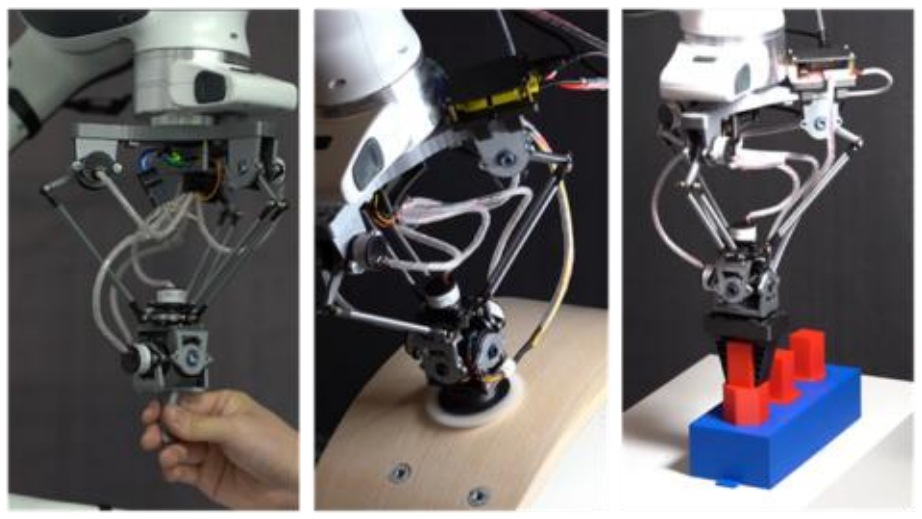

| Scheme | Cost | Build Effort | Bandwidth | Accuracy | Compliance |
|---|---|---|---|---|---|
| **Delta6(Ours)** | Low | Easy | Medium | Good | Compliant |
| ATI Nano | Very high | Factory | Ultra-high | Excellent | Rigid |
| Capacitive [11] | Moderate | Clean-room | n/a | Good | Rigid |
| Piezoresistive [12] | Moderate | Clean-room | High | Good | Rigid |
| Photo-interrupters [13] | Very low | Moderate | High | Basic | Low |
| Optoelectronic [14] | Very low | Moderate | Medium | Basic | Low |
| FBG multilayer [15] | Very high | Difficult | Medium | Excellent | Moderate |
| Fiber-optic [16] | Very low | Difficult | Medium | Good | Low |
| Vision fiducial [17] | Low | Easy | Very low | Basic | Compliant |

**Fig. 1.** Delta6 Prototype in Action and Qualitative Benchmark Against Representative 6-DOF Force/Torque Sensors (disadvantages highlighted in red).

inherently compliant, and long-term-robust alternatives can significantly broaden the adoption of force-sensing solutions.

This paper introduces **Delta6**, a spring-driven, magnetic-encoder based 6-DOF F/T sensing robot end-effector. A complete prototype, including a US$249 single-board computer, costs about US$583 in total yet can be reduced further through design iterations and economies of scale; each component is either off-the-shelf or can be 3D printed, and the assembly tolerances are no stricter than those of flat-pack furniture. The first prototype handles up to 25.1 N force-norm and 0.58 N·m torque-norm. Using our parametric design model, we further demonstrate several variants that extend the load range well beyond these baseline limits. After removing the top 1% of outliers, the uncalibrated sensor shows per-axis errors below 7 % of full scale (FS). Training a lightweight machine-learning model on a single dataset further reduces these per-axis errors to 3.8 % FS while preserving a 50 Hz inference rate. Its inherent compliance makes this bandwidth sufficient for impedance-based force control, enabling contact-rich tasks such as wood buffing and tight-clearance peg-in-hole insertion. A detailed qualitative comparison with previously studied schemes appears in Fig. 1.

The main contributions are summarized as follows:

- *Modular passive sensor:* A modular, low-cost, and inherently compliant 6-DOF force/torque sensing end-effector is introduced, integrating a classical delta translational stage with antagonistic torsion springs and magnetic encoders. All parts are off-the-shelf or 3D printed, making the device easy to replicate, maintain, and scale.
- *Analytical & learned estimation:* A quasi-static analytical model supports parametric design and provides an MCU-deployable baseline, while sequence models (LSTM [19], GRU [20], Transformer [21]) serve as practical error-compensation layers for non-idealities. Maximum update rates are benchmarked on multiple platforms. Accuracy, dynamic response, and robustness are verified through a series of tests.
- *Impedance-controlled demos:* Integrated with the proposed hybrid force–position Cartesian-impedance controller, Delta6 successfully executes two contact-rich tasks, buffing an unmodelled curved surface and inserting a peg with 0.2 mm clearance, demonstrating its practicality.

The remainder of this paper is organized as follows. Section II introduces the overall mechanical design of Delta6, including the parallel structure and spring arrangement. Section III details the modeling and sensing algorithms, covering analytical and learning-based methods. Section IV models the impedance controller and applications task design. Section V presents the experimental setup and validation results, conducts extensive benchmarks, and discusses the insights gained. Section VI summarizes the findings and concludes the paper.

## II. MECHANICAL DESIGN

### A. Overall Structure

Fig. 2 presents an exploded view of the Delta6 assembly, based on a well-established delta robot kinematic configuration, providing three translational degrees of freedom (DOFs) along the $x$, $y$, and $z$ axes [22-24]. Additionally, a compliant 3-DOF antagonistic spring unit (Fig. 2-d) introduces three rotational DOFs, achieving full 6-DOF motion at the Tool Center Point (TCP).

The manipulator comprises three identical 1-DOF antagonistic spring units (Fig. 2-b), symmetrically arranged on the base (Fig. 2-a), each rotated by 120 degrees around the central axis with an inclination of 30 degrees relative to the horizontal plane. These units utilize torsion springs arranged antagonistically to produce balanced opposing torque, measured precisely by integrated magnetic encoders.

Connecting each 1-DOF unit to the central 3-DOF antagonistic spring unit are parallelogram linkages (Fig. 2-c). Each linkage pair consists of two rods with ball joints at both ends, forming stable parallelograms. Rotational DOFs about each rod's axis are minor and thus considered negligible, simplifying the assembly process.

### C. 1-DOF Antagonistic Spring Unit

As illustrated in Fig. 2, each 1-DOF antagonistic-spring module consists of a 3D-printed proximal link with an integrated shaft, two bearing housings, two bearings, a magnetic encoder and two torsion springs, together forming a compliant revolute joint. The two springs are mounted in opposite directions, and the angle of static torque equilibrium can be fine-tuned by adjusting the spring clips. Under external loading, the spring torques superpose, thereby increasing joint stiffness and mechanical stability, eliminating backlash, and relaxing assembly-precision requirements.

### D. 3-DOF antagonistic spring unit

The core component of 3-DOF antagonistic spring unit (Fig. 2-d) is the cross-shaped two-way shaft, which together with two sets of bearings, forms a two-degree-of-freedom universal joint capable of rotating about its center. By incorporating two sets of torsional springs, identical to those used in the 1-DOF Antagonistic Spring Unit, this universal joint is made compliant. Structurally, this design can be considered a two-dimensional extension of the previously described 1-DOF Antagonistic Spring Unit.

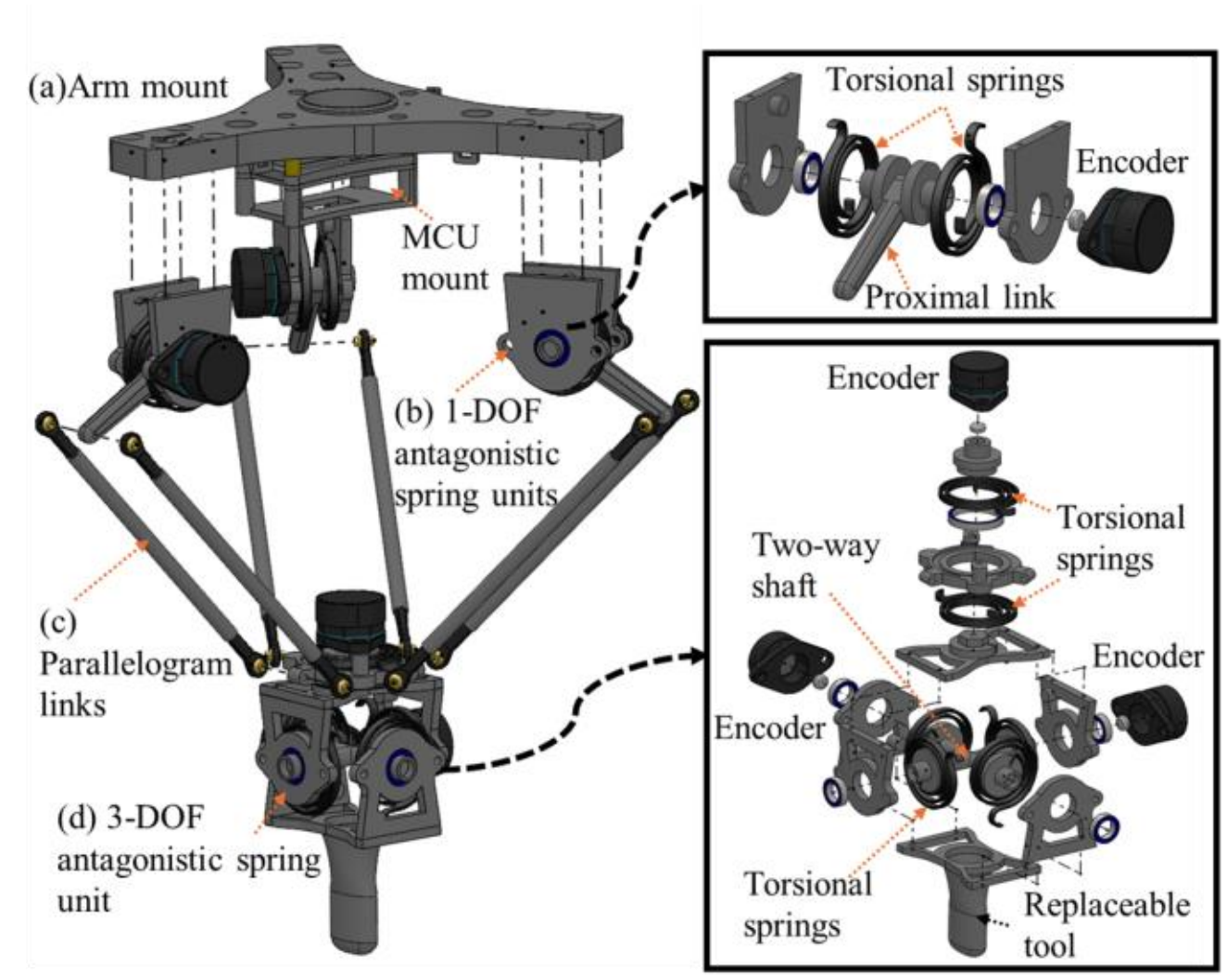

**Fig. 2.** Exploded View Diagram of the Delta6 Assembly. (a) Mounting structure for the robotic arm, including a substructure designed for securing the microcontroller.(b) 1-DOF antagonistic spring unit, including proximal link, with a total of three identical units. (c) Parallelogram links, consisting of six rods arranged in three pairs, each rod featuring a ball joint at both ends. (d) 3-DOF antagonistic spring unit, enabling movement along three rotational degrees of freedom at the end.

The z-axis design differs from the $x$- and $y$-axes, as it must account for axial forces in addition to rotational forces. However, the underlying principle remains consistent with the other joints: flexibility is achieved through an antagonistic spring configuration.

Additionally, a replaceable grip handle, depicted in the diagram, facilitates manual operation and testing, and can readily be replaced with specialized tools like grippers or grinders.

## III. MODELING AND SENSING ALGORITHMS

### *A. Definition of Structural Parameters*

Fig. 3-a depicts Delta6 in its neutral pose, where the 6 joint angles are denoted as $\theta_i\ (i = 1, \dots, 6)$ and their positive rotation directions are indicated. Several key coordinate frames are introduced to describe the system kinematics. The coordinate frame $\{M\}$ represents the endpoint of the last link of the robotic arm, where the Delta6 is mounted. The coordinate frame $\{B\}$ serves as the base frame of the Delta robot configuration. It is located at the center of the triangle formed by the axes of the three 1-DOF joints. In this frame, the $x$-axis is perpendicular to the axis of the first joint, while the $z$-axis is aligned with the normal of the triangular plane. The transformation from $\{M\}$ to $\{B\}$ is not fixed, as it depends on how the robotic arm's kinematics are defined and how the Delta6 is mounted. The origin of frame $\{E\}$ is located at the center of the 3-DOF joint and rigidly attached to the final linkage, giving it a fixed transformation with respect to the TCP frame $\{T\}$. In the neutral configuration, frame $\{E\}$'s $x$-, $y$-, and $z$-axes are respectively coaxial with joints 4, 5, and 6; we denote this neutral frame as $\{E_0\}$. Delta6 aims to measure the 6-DOF wrench at frame $\{E\}$, meaning that the measured forces/torques correspond to the components acting at this frame.

As shown in Fig. 3-b, the $x$-axis of frame $\{B\}$ points toward joint 1. Joint 2 is obtained by rotating joint 1 counterclockwise around the $z$-axis of $\{B\}$ by an angle $\varphi_2$, while joint 3 is obtained by rotating joint 1 counterclockwise by $\varphi_3$. As shown in Fig. 3-a. In this design, $\varphi_2 = \varphi_3 = 120^{\circ}$, making the overall structure three-fold centrally symmetric. Furthermore, three additional coordinate frames $\{B_1\}$, $\{B_2\}$ and $\{B_3\}$, are defined, where the frame $\{B_1\}$ coincides completely with $\{B\}$, while $\{B_2\}$ and $\{B_3\}$ are obtained by rotating $\{B\}$ counterclockwise around its $z$-axis by angles $\varphi_2$ and $\varphi_3$, respectively.

As shown in Fig. 3-c, the angles $\theta_i\ (i = 1, 2, 3)$ are defined as the angles between the $x$-axis of their respective frames $\{B_i\}$ and the corresponding joint axes. As shown in Fig. 3-a, $\theta_4$, $\theta_5$ and $\theta_6$ represent the rotational displacements of the three components of the 3-DOF joint. They are defined so that, when there is no rotation in the transformation from $\{B\}$ to $\{E\}$, $\theta_4 = \theta_5 = \theta_6 = 0^{\circ}$. When external forces cause deformation in the joints, the *ZXY* intrinsic Euler angles from $\{B\}$ to $\{E\}$, corresponds to $\theta_6$, $\theta_4$ and $\theta_5$. Besides, the encoder reading for displacement is defined as $\theta_{ei}$. For $\theta_{ei} (i = 1, 2, 3)$, it is given by $\theta_{ei} = \theta_i - \theta_{offset}$, where $\theta_{offset}$ represents the natural angle of the proximal link, i.e., the angle when the torque is zero. In this design, $\theta_{offset} = 30^{\circ}$. For $\theta_{ei} (i = 4, 5, 6)$, the encoder reading is directly $\theta_{ei} = \theta_i$.

Moreover, as shown in Fig. 3-a, the Delta robot structure is defined by several key design parameters, $a$, $l_a$, $l_b$ and $b$. The parameter $a$ represents the distance from the coordinate system center to the three joint axes, while $l_a$ and $l_b$ denote the lengths of the proximal links and the parallelogram linkages, respectively. The parameter $b$ is determined by extending the axes of the three joints on the 3-DOF joint that connect to the parallelogram linkages, forming an equilateral triangle parallel to the base triangle. The center of this equilateral triangle, denoted as point $P$, is equidistant from the centers of the three joints, with this distance denoted as $b$. By translating this center along the $z$-axis of $\{B\}$ by a distance $c$, the origin of the $\{E\}$ coordinate frame is defined.

### *B. Analytical Approach*

Delta6 measures the 6-DOF forces and torques at the $\{E\}$ coordinate frame using encoder readings. The system takes $\theta_{ei}\ (i = 1, \dots, 6)$ as inputs and computes the resulting external wrench $\boldsymbol{W}$. This section presents a quasi-static wrench estimation method based on static equilibrium, intended as a transparent baseline and a parametric design tool. The formulation neglects dynamic effects and does not explicitly model friction/backlash or structural compliance.

The forward kinematics (FK) of the Delta robot, which maps the joint angles $\theta_1$, $\theta_2$ and $\theta_3$ to the Cartesian coordinates of point ${}^{B}\boldsymbol{P} = [p_x,\ p_y,\ p_z]^{\top}$ is well documented in the literature [24].

The vector $\boldsymbol{v}_i\ (i = 1, 2, 3)$ represents the midline of the parallelogram linkage of the $i^{th}$ branch in $\{B\}$, as shown in Fig. 3-c, can be obtained by:

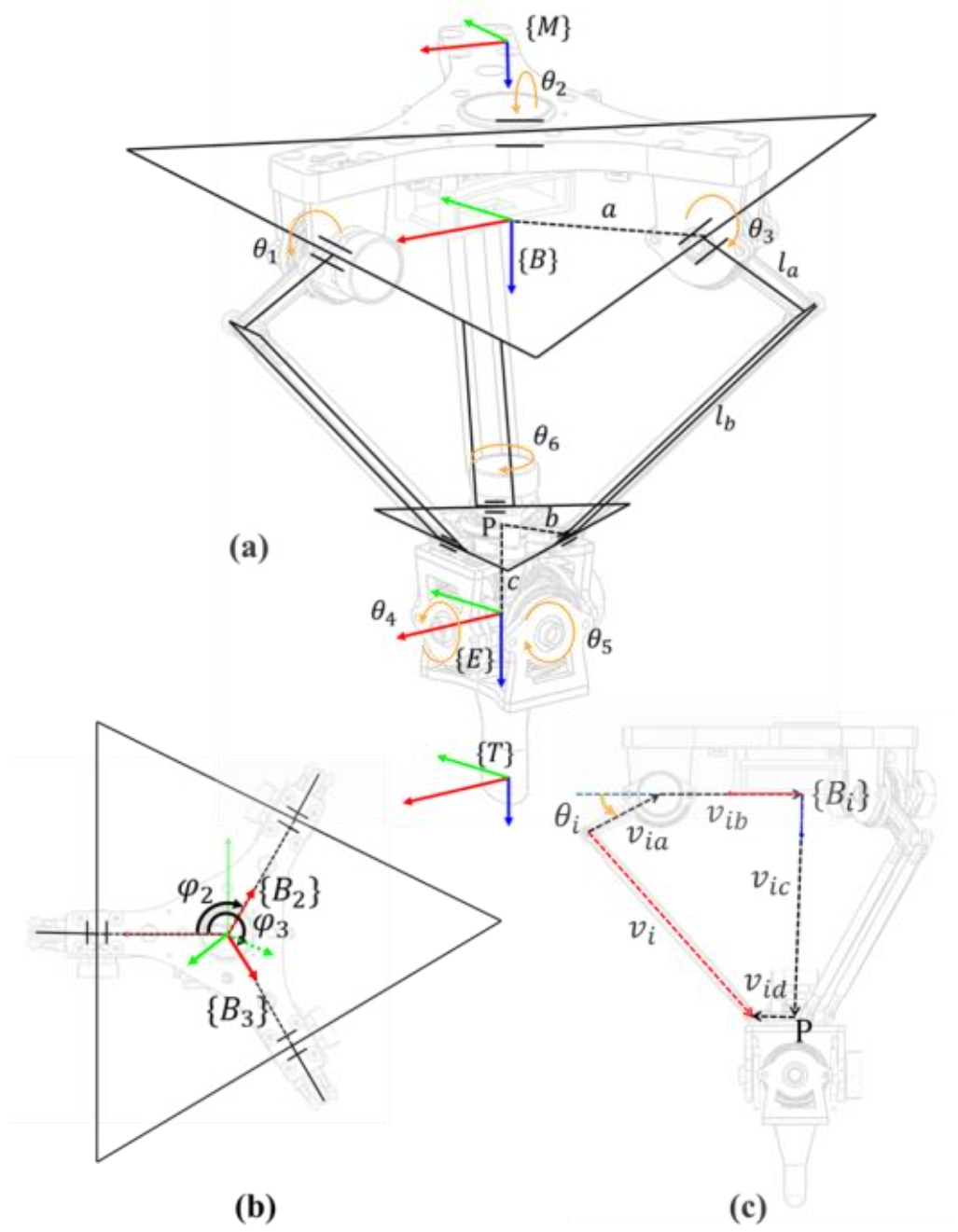

**Fig. 3.** Structural Parameters Definition of Delta6. (a) 3D View, defines the key coordinate positions, Delta6 design parameters, and the indices & orientations of the six encoders. (b) Top View, shows the angle notation between different branches. $\{B_2\}$ and $\{B_3\}$ are obtained by rotating $\{B\}$ counterclockwise around the z-axis by angles $\varphi_2$ and $\varphi_3$, respectively. (c) Side View of Branch $i$, highlights the key vectors that are essential for force calculations.

$$\boldsymbol{v}_i = \boldsymbol{v}_{ia} + \boldsymbol{v}_{ib} + \boldsymbol{v}_{ic} + \boldsymbol{v}_{id} \tag{1}$$

where

$$\boldsymbol{v}_{ia} = [-l_a \cos\theta_i \cos\varphi_i\,, -l_a \cos\theta_i \sin\varphi_i\,, l_a \sin\theta_i]^\top$$
$$\boldsymbol{v}_{ib} = [-a\cos\varphi_i\,, -a\sin\varphi_i\,, 0]^\top$$
$$\boldsymbol{v}_{ic} = [p_x,\ p_y,\ p_z]^\top$$
$$\boldsymbol{v}_{id} = [b\cos\varphi_i\,, b\sin\varphi_i\,, 0]^\top.$$

Therefore,

$$\boldsymbol{v}_i = \begin{bmatrix} p_x - l_a \cos\theta_i \cos\varphi_i + (b-a)\cos\varphi_i \\ p_y - l_a \cos\theta_i \sin\varphi_i + (b-a)\sin\varphi_i \\ p_z - l_a \sin\theta_i \end{bmatrix}. \tag{2}$$

Because the length of parallelogram linkages equal to $l_b$, the unit vector of $\boldsymbol{v}_i$ can be obtained by equation:

$$\hat{\boldsymbol{v}}_i = \frac{\boldsymbol{v}_i}{\|\boldsymbol{v}_i\|} = \frac{\boldsymbol{v}_i}{l_b} \tag{3}$$

Then, the unit vector $\hat{\boldsymbol{v}}_i$ can be expressed in each branch coordinate frame $\{B_i\}$ using the following equation:

$$^{B_i}\hat{\boldsymbol{v}}_i = {}^{B_i}_{B}\boldsymbol{R} \cdot \hat{\boldsymbol{v}}_i = \left[\hat{v}_{i,x}, \hat{v}_{i,y}, \hat{v}_{i,z}\right]^\top, \tag{4}$$

where ${}^{B_i}_{B}\boldsymbol{R} \in SO(3)$ is the rotational matrix form frame $\{B_i\}$ to $\{B\}$, which denoted as

$${}^{B_i}_{B}\boldsymbol{R} = \boldsymbol{R}_z(-\varphi_i). \tag{5}$$

Since the force ${}^{B_i}\boldsymbol{F}_i$ exerted on each proximal link by parallelogram link is directed along ${}^{B_i}\hat{\boldsymbol{v}}_i$, the torques $\boldsymbol{\tau}_i$ $(i = 1, 2, 3)$ generated at joints 1, 2, 3 follow the relationship:

$$\boldsymbol{\tau}_i = \boldsymbol{r} \times {}^{B_i}\boldsymbol{F}_i, \tag{6}$$

where

$$\boldsymbol{r} = [l_a \cos\theta_i\,, 0, l_a \sin\theta_i],$$
$${}^{B_i}\boldsymbol{F}_{i_i} = \|\boldsymbol{F}_i\|\,{}^{B_i}\hat{\boldsymbol{v}}_i\,.$$

Expanding the cross product above, the magnitude of each applied force can be obtained:

$$\|\boldsymbol{F}_i\| = \frac{\tau_{i,y}}{l_a(\hat{v}_{i,x}\sin\theta_i - \hat{v}_{i,z}\cos\theta_i)}. \tag{7}$$

By knowing the spring unit coefficient k, it has:

$$\tau_{i,y} = \tau_i = \mathrm{k}\theta_{ei}. \tag{8}$$

Then, the expression of each force in the $\{B\}$ coordinate system can be determined:

$$\boldsymbol{F}_i = \|\boldsymbol{F}_i\|\hat{\boldsymbol{v}}_i. \tag{9}$$

By summing up the forces, we obtain the force at point $P$, which is equivalent to the force acting on frame $\{E_0\}$. Since we are calculating the reaction force, a negative sign is applied:

$${}^{E_0}\boldsymbol{F} = -(\boldsymbol{F}_1 + \boldsymbol{F}_2 + \boldsymbol{F}_3) = \left[{}^{E_0}F_x,\ {}^{E_0}F_y,\ {}^{E_0}F_z\right]^\top. \tag{10}$$

Furthermore, the torque ${}^{E_0}\boldsymbol{M}$ acting on frame $\{E_0\}$, is equal to the torque on joints 4, 5, and 6. Thus, it has:

$${}^{E_0}\boldsymbol{M} = [{}^{E_0}M_x, {}^{E_0}M_y, {}^{E_0}M_z]^\top = [k\theta_{e4}, k\theta_{e5}, k\theta_{e6}]^\top. \tag{11}$$

To represent the wrench at frame $\{E\}$, it has

$${}^{E}\boldsymbol{F} = {}^{E_0}_{E}\boldsymbol{R}^\top \cdot {}^{E_0}\boldsymbol{F} = \left[F_x,\ F_y,\ F_z\right]^\top,$$
$${}^{E}\boldsymbol{M} = {}^{E_0}_{E}\boldsymbol{R}^\top \cdot {}^{E_0}\boldsymbol{M} = \left[M_x,\ M_y,\ M_z\right]^\top, \tag{12}$$

where ${}^{E_0}_{E}\boldsymbol{R} \in SO(3)$ is the rotational matrix form frame $\{E_0\}$ to $\{E\}$, which can be calculate from $\theta_{e4}, \theta_{e5}, \theta_{e6}$. Therefore, we have the final expressions of external wrench acting on the frame $\{E\}$:

$$\boldsymbol{W}(\boldsymbol{\theta}_e) = [F_x, F_y, F_z, M_x, M_y, M_z]^\top. \tag{13}$$

### *C. Numerical Methods*

Accurately estimating end-effector forces with a purely analytical model can be hindered by real-world factors, especially when cost-driven design compromises are considered. These factors mainly include:

- Hysteresis in compliant elements.
- Limited rigidity of 3D-printed parts.
- Ball-joint friction and backlash.
- Assembly imperfections.

To mitigate these non-idealities in a practical manner, we adopt an end-to-end, data-driven compensation strategy. Encoder–wrench data are collected and used to learn a direct mapping to the external wrench $\boldsymbol{W}(\boldsymbol{\theta}_e)$ with time-series networks. Numerical force–torque estimation methods have been explored in prior works [25–27]; here we use sequence models as an error-compensation layer and focus on deployment-relevant accuracy–latency trade-offs and reproducible empirical evaluation.

To assess the inductive biases of different sequence models, we train and evaluate GRU, LSTM, and Transformer networks on a common dataset. GRU, as the simplest gated recurrent unit, uses minimal gating to capture short to mid-range dependencies, allowing us to test whether a lightweight architecture is sufficient to compensate for hysteresis and backlash. LSTM, the standard gated RNN, stores information in a cell state and controls it with an output gate, allowing it to capture fast friction transients and slower material creep. Transformer leverages parallel self-attention to attend to all joints and time steps within each input window, enabling the identification of cross-joint, cross-time couplings. For example, the phase shift when deflection in one link induces micro-oscillations in neighboring joints, or the shape difference between loading and unloading hysteresis loops. By attending to all time steps without recursive state propagation, self-attention can, in principle, help model pronounced hysteresis and global load redistribution in compliant mechanisms [28–29]. All three networks are universal function approximators with temporal memory and therefore have the potential to absorb hysteresis, compliance, friction, backlash, and assembly-related nonlinear effects that depend on motion history.

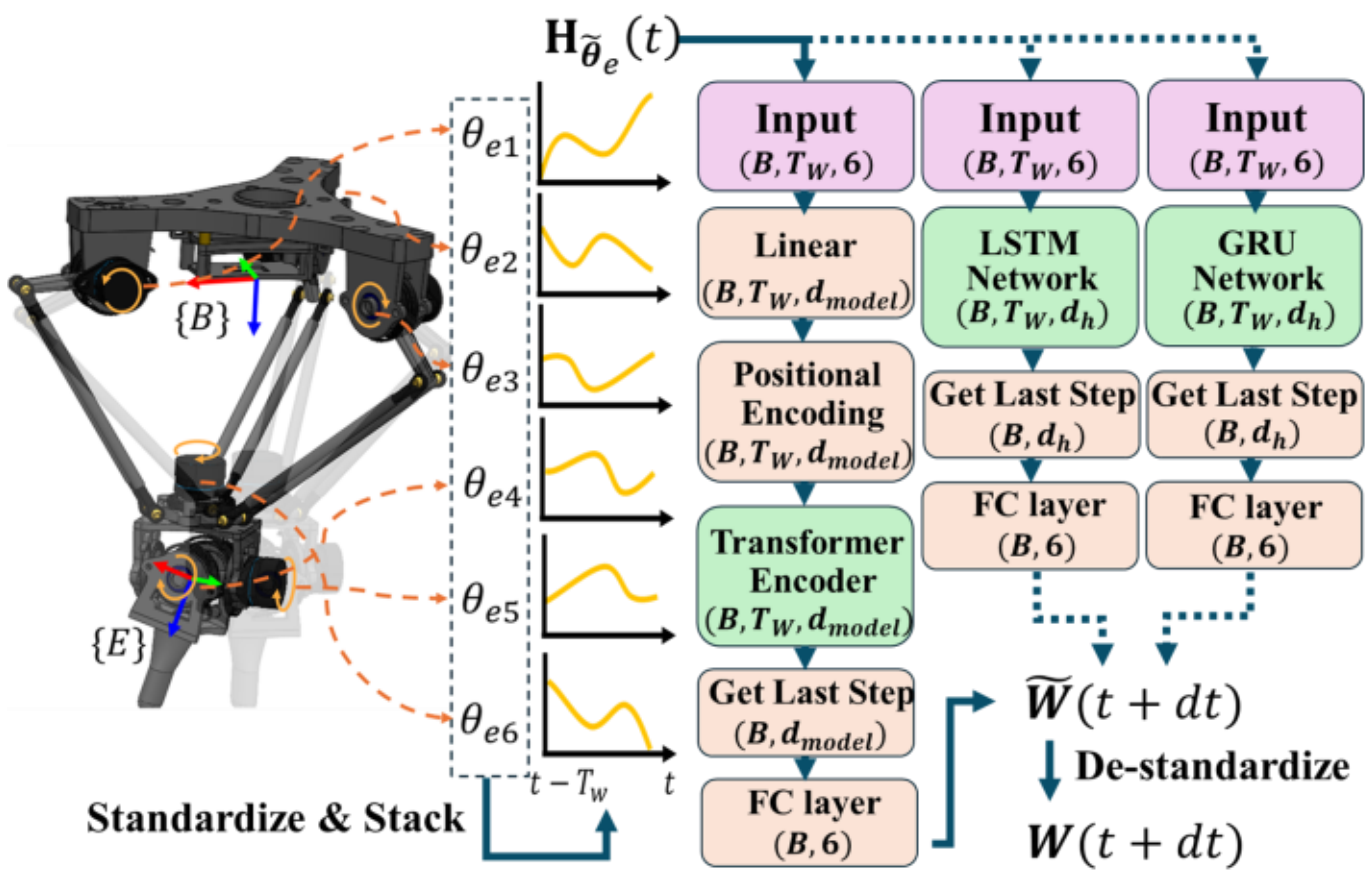

**Fig. 4.** Architectures for encoder-to-wrench estimation. Encoder histories $\mathbf{H}_{\tilde{\theta}_e}(t)$ enters a Transformer (left), LSTM (centre), or GRU (right) to predict the six-dimensional wrench. During inference the batch dimension is $B = 1$.

*1)Data collection*

Assuming a 6-DOF force/torque sensor is installed at $\{T\}$, the key data to be collected include: $t$ , $\boldsymbol{\theta}_e$, $\boldsymbol{W}$, where $t$ marks the time stamps, $\boldsymbol{\theta}_e$ are the encoder reading of all six joints, while $\boldsymbol{W}$ is the wrench acting on $\{E\}$.

Among these, $\boldsymbol{W}$ is not directly available, as the sensor records data in frame $\{T\}$, denoted as ${}^{T}\boldsymbol{W}$ where obtains ${}^{T}\boldsymbol{F} = [{}^{T}F_x,\ {}^{T}F_y,\ {}^{T}F_z]^\top$, and ${}^{T}\boldsymbol{M} = [{}^{T}M_x,\ {}^{T}M_y,\ {}^{T}M_z]^\top$. In order to obtain $\boldsymbol{W}$, the following formula is applied:

$$
\begin{aligned}
{}^{E}\boldsymbol{F} &= {}^{T}_{E}\boldsymbol{R} \cdot {}^{T}\boldsymbol{F}, \\
{}^{E}\boldsymbol{M} &= {}^{T}_{E}\boldsymbol{R} \cdot ({}^{T}\boldsymbol{M} - {}^{T}_{E}\boldsymbol{p} \times {}^{T}\boldsymbol{F}).
\end{aligned} \tag{14}
$$

Where ${}^{T}_{E}\boldsymbol{R} \in SO(3)$ is a constant rotational matrix, from frame $\{T\}$ to $\{E\}$, defined by the rigid transform of the terminal link. ${}^{T}_{E}\boldsymbol{p} \in \mathbb{R}^3$ is the corresponding translation.

*2) End-to-End Machine Learning Methods*

A sequence of six $\boldsymbol{\theta}_e$ encoder readings serve as the model input, while the measured $\boldsymbol{W}$ values are used as labels for supervised learning. To ensure numerical stability, all inputs and outputs are first standardized. As shown in Fig. 4, a sliding window of length $T_w$ is applied to construct sequential scaled inputs $\mathbf{H}_{\tilde{\theta}_e}(t) = \{\widetilde{\boldsymbol{\theta}}_e(t - T_w), \dots, \widetilde{\boldsymbol{\theta}}_e(t)\}$, to predict scaled wrench $\widetilde{\boldsymbol{W}}$ at $t + dt$, allowing the model to incorporate temporal dependencies. Therefore, the input of model can also be denoted as $\boldsymbol{X} \in \boldsymbol{R}^{B \times T_w \times n_{in}}$, while the output denoted as $\boldsymbol{Y} \in \boldsymbol{R}^{B \times n_{out}}$, where $B$ is batch size; $n_{in}$ and $n_{out}$ are input/output sizes that equal to 6.

During training and inference, the Transformer model first maps each input time step into a feature space with a dimensionality equal to the model dimension $d_{model}$, incorporating a learnable positional encoding to retain temporal ordering. The encoded sequence is then processed through multiple layers of self-attention and feedforward networks. Finally, the output layer extracts the hidden state of the last time step to generate the predicted torque and force.

For the LSTM, multiple layers of long short-term memory units are stacked to handle temporal dependencies, with the hidden size $d_h$ governing the dimensionality of the recurrent hidden state. At each time step, the hidden state is updated based on both the current input and the previous hidden state, allowing the model to accumulate information across the entire sequence. In the final step, the last hidden state is mapped through a fully connected layer to produce the torque or force predictions.

Similarly, GRU replaces the LSTM cell with a gated recurrent unit that relies on update and reset gates, reducing parameter count while still preserving temporal context.

Cross-validation and early stop techniques are needed to optimize model performance. For the Transformer model, the hyperparameters include model dimension $d_{model}$, number of attention heads $n_h$, number of layers $N_L$ and dropout rate $p_{drop}$. For LSTM/GRU, the key hyperparameters are the hidden size $d_h$, the number of layers $N_L$, the dropout rate $p_{drop}$ as well.

## IV. APPLICATIONS VALIDATION

This section first demonstrates how the Delta6, when integrated with a 6-DOF manipulator, realizes hybrid force–position regulation by overlaying a Cartesian-impedance layer at the TCP. To assess the robustness of this controller and to quantify Delta6's practical capabilities, two tasks were designed. The first is force-regulated wood buffing on an unmodelled curved surface, selected to verify that Delta6 maintains stable performance under the high-frequency vibrations condition of polishing processes. The second is force-guided peg-in-hole insertion with sub-millimeters clearance, intended to evaluate Delta6's ability to sense and modulate fine contact forces during precision assembly.

### *A. Hybrid Force-Impedance Control*

The translational part of the transform from frame$\{B\}$ to $\{E\}$ is ${}^{B}_{E}\boldsymbol{p} = [p_x,\ p_y,\ p_z + c]^\top$, whereas the rotational part ${}^{B}_{E}\boldsymbol{R}$ is constructed from the *ZXY* intrinsic Euler angles $\theta_{e6}, \theta_{e4}, \theta_{e5}$; together, they form the homogeneous matrix ${}^{B}_{E}\boldsymbol{T}$.

The mounting transform ${}^{M}_{B}\boldsymbol{T}$ and tool transform ${}^{E}_{T}\boldsymbol{T}$ are constant once the hardware is assembled. Thus, the TCP pose relative to the flange is obtained by ${}^{M}_{T}\boldsymbol{T} = {}^{M}_{B}\boldsymbol{T}\,{}^{B}_{E}\boldsymbol{T}\,{}^{E}_{T}\boldsymbol{T}$. Stacking its translation and intrinsic *XYZ* Euler angles gives the six-vector ${}^{M}_{T}\boldsymbol{X} \in \boldsymbol{R}^{6\times1}$. When the robot arm system reports the global flange pose ${}^{G}_{M}\boldsymbol{T}$, the TCP pose in the global frame $\{G\}$ is ${}^{G}_{T}\boldsymbol{T} = {}^{G}_{M}\boldsymbol{T}\,{}^{M}_{T}\boldsymbol{T}$, whose 6-vector form is denoted as ${}^{G}_{T}\boldsymbol{X}$ or $\boldsymbol{X}$.

The desired TCP behavior is enforced through six decoupled mass-spring-damper equations,

$$
\boldsymbol{M}\ddot{\boldsymbol{X}} + \boldsymbol{B}\dot{\boldsymbol{X}} + \boldsymbol{K}(\boldsymbol{X} - \boldsymbol{X}_d) = {}^{T}\boldsymbol{W}. \tag{15}
$$

which realizes a Cartesian impedance law. Here $\boldsymbol{M}$, $\boldsymbol{B}$ and $\boldsymbol{K} \in \boldsymbol{R}^{6\times6}$, are diagonal matrices that define the virtual inertia, damping and stiffness perceived at the TCP. $\boldsymbol{X}_d$ is an operator-defined set-point that can be updated online to command motion while preserving the impedance profile. The external wrench $^{T}\boldsymbol{W}$ is obtained by transforming the raw Delta6 measurement $\boldsymbol{W}$ into the TCP frame using the inverse wrench-transformation described in (14). Solving (14) at each control cycle yields $(\ddot{\boldsymbol{X}}, \dot{\boldsymbol{X}}, \boldsymbol{X})$; The 6-vector $\boldsymbol{X}$ is then re-packed into the homogeneous transform $^{G}_{T}\boldsymbol{T}$ . Multiplying by the inverse of the tool matrix gives the flange pose $^{G}_{M}\boldsymbol{T} = {}^{G}_{T}\boldsymbol{T}({}^{M}_{T}\boldsymbol{T})^{-1}$, which is then transformed back to 6-vector $^{G}_{M}\boldsymbol{X}$ and transmitted to the robot controller as flange-level motion commands.

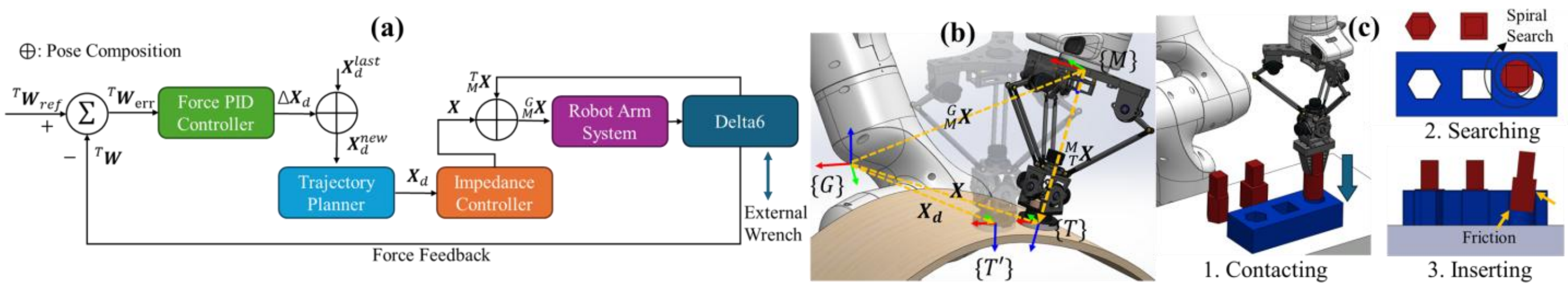

**Fig. 5.** Integrated Force-Control Applications with Delta6. a) Hybrid Force-Impedance Control Architecture. b) Key Pose Frames during Curved-Wood Buffing. c) Three-Phase Peg-in-Hole Assembly Task

To enable precise force regulation at TCP, we augment the existing impedance controller with a PID-based force-feedback loop (see Fig. 5-a). The system continuously measures $^{T}\boldsymbol{W}$ and compares it with the desired wrench $^{T}\boldsymbol{W}_{ref}$. The resulting error is processed by the PID controller and converted into a pose increment vector $\Delta\boldsymbol{X}_d$. This increment is applied to the previous desired pose $\boldsymbol{X}_d^{last}$ to obtain the new desired pose $\boldsymbol{X}_d^{new}$.

The updated pose $\boldsymbol{X}_d^{new}$ is then submitted to the trajectory planner, which embeds the task-level reference path into the pose stream and generates the next $\boldsymbol{X}_d$ for the impedance controller, thereby realizing hybrid force/position.

Overall, any external disturbance sensed by Delta6 is fed immediately into the PID loop, allowing the controller to retarget the robot flange in real time and maintain the commanded interaction force. Meanwhile, the end-effector's own elastic deflection $^{M}_{T}\boldsymbol{X}$ is incorporated into the impedance-control output, refining the flange command. As a result, the TCP presents itself to the outside world as a stiffness–damping–mass-controllable tool whose motion is modulated by a blend of position and force regulation.

### *B. Wood Buffing on an unknown-geometry Curved Surface*

The set-up for this task is illustrated on the right-hand side of Fig. 5-b. Delta6 is mounted at the wrist of a 6-DOF manipulator and fitted with a rotary grinding head. The operation is divided into two sequential phases, *Contacting* and *Polishing*, both executed under the hybrid force–impedance controller.

*Contacting phase* - The robot is first positioned a small offset above the curved wooden workpiece while the grinder spindle is brought up to speed. The trajectory planner then commands the desired TCP pose $\boldsymbol{X}_d$ to descend along the $z$-axis. As soon as the measured normal force $F_z$ exceeds a predefined threshold, control transitions automatically to the polishing phase.

*Polishing phase* - A dedicated path generator produces a zig–zag scan pattern in the $x - y$ plane. At every control tick the trajectory planner overwrites the $x-$ and $y-$ components of the force control output $\boldsymbol{X}_d^{new}$ , while leaving the remaining unchanged. The wrench reference $^{T}\boldsymbol{W}_{ref}$ is configured to only impose a constant downward force in $z$. Because references of $M_x$ , $M_y$ are zero, the force loop is expected to adjust the robot so that the TCP remains normal to the unknown surface while maintaining the specified normal force. The task is successful once all waypoints of the zig–zag trajectory have been executed without violating the force or torque limits.

### *C. Force/Torque Control for Tight Assembly*

This task employs the same hybrid force–position impedance controller to drive a 6DOF robot equipped with the Delta6 and a gripper for tight peg-in-hole assembly. As shown in Fig. 5-c, it proceeds through three phases:

*Contacting* - The trajectory planner lowers the TCP along the $z$-axis until a small normal force is detected, thereby establishing compliant rim contact.

*Searching* - while holding the preload constant, the planner executes an outward spiral in the $x - y$ plane, superimposed with a mild yaw dither, allowing non-circular pegs to self-align with the bore axis under active compliance.

*Inserting* - Once the normal force drops, signaling entry into the bore, the 6-DOF force control loop keeps the tool-frame wrench $^{T}\boldsymbol{W}$ near zero and advances by micro-steps under zero-torque constraints, driving the peg to depth without breaching the global wrench limits.

In addition, to introduce a graded level of difficulty, three peg–hole geometries - cylindrical, square, and hexagonal - are evaluated separately, and the insertion success rate will be recorded for each case.

## V. EXPERIMENTS

### *A. Prototype Implementation*

#### *1) Materials and Deployment*

Table I presents the list of materials used in the first prototype of Delta6 along with their estimated prices. All customized components were fabricated by using a consumer-grade 3D printer with PLA filament. while torsion springs and the ball joints–bearing sets were sourced from commodity doorhandle and RC-hobby suppliers, respectively, ensuring low unit cost and wide availability. The only required high-cost item is the ERCK-05 SPI 14-bit magnetic encoder; less expensive alternatives could be adopted in later revisions. If no on-board computing resource is available, a Jetson Nano Super module

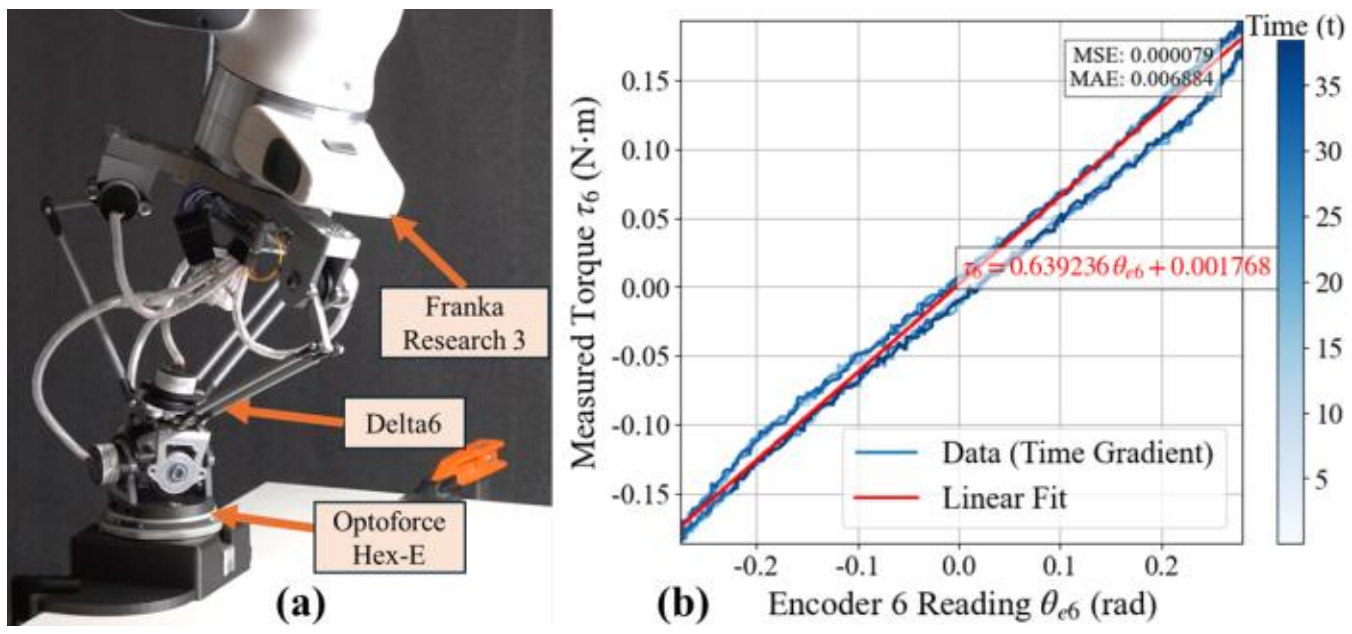

**Fig. 6.** a) Experimental setup for automated acquisition of wrench–encoder synchronised samples. b) Spring Constant from Encoder 6 Reading vs. Measured Torque.

(US$249) may be incorporated to execute learning-based force-inference algorithms; this remains optional, as Arduino Nano Every already supports the analytical force solution at sampling rates of at least 50 Hz (see Section B for benchmarks). Excluding the optional single-board compute module, the material cost of the prototype is US$334 at retail, and volume production is expected to reduce the total bill of materials to below US$100. Assembly, dominated by wiring and soldering, requires approximately two hours; adopting a dedicated printed-circuit board would significantly shorten this time and further lower associated labor costs.

TABLE I
MATERIAL AND COST ESTIMATION

| Material | Quantity | Estimated Unit Cost (USD) |
|---|---|---|
| PLA Filament (200 g) | 1 | 3.00 |
| Arduino Nano Every | 1 | 13.00 |
| M2 Tie Rod Ball End | 12 | 1.00 |
| M2×100 mm Threaded Rod | 6 | 2.00 |
| Manganese Steel Door Handle Spring | 12 | 2.00 |
| ERCK 05SPI 14-bit Magnetic Encoder | 6 | 43.00 |
| 15×21×4 mm Ball Bearing | 1 | 1.00 |
| 8×12×3.5 mm Ball Bearing | 10 | 1.00 |
| Wires, Connectors, and Fasteners | 1 | 1.00 |
| *Jetson Nano Super (optional)* | 1 | 249.00 |
| **Subtotal (Without Jetson)** | | **334.00** |
| **Total** | | **583.00** |

The microcontroller used in this prototype is the Arduino Nano Every, which functions as an SPI interface for reading encoder data and transmitting it to a PC via serial communication. Except for inference time benchmark, all experiments were conducted on a desktop (Intel i7-12700K CPU, 2080ti GPU).

As illustrated in Fig. 6-a, the Delta 6 prototype was mounted between a Franka Research 3 (FR3) collaborative manipulator and an OptoForce HEX-E 6-axis F/T sensor. The sensor's own measurement frame was designated $\{T\}$, and its wrench output was mapped into the Delta 6 base frame $\{E\}$, via formula (14). The FR3 served as the active side of the pair. Cartesian flange poses were both polled and commanded through WinGs Operating Studio (WOS) middleware [30]. This configuration enabled automated collection of synchronized wrench–encoder-angle pairs for training and evaluation.

*2) Spring Unit Characterization*

The Delta6 prototype is equipped with six elastic units, each consisting of a pair of identical antagonistic springs. To characterize the mechanical properties of a single elastic unit, joint 6 was selected for measurement, allowing the determination of the elastic coefficient $k$.

A total of 1,998 data points $(t, \theta_{e6}, \tau_6)$, were collected at 50 Hz, corresponding to approximately 40 seconds of data acquisition. Only joint 6 was driven through repeated ± rotations during this collection. As shown in Fig. 6-b, a linear regression was then applied to the collected data, yielding an estimated elastic coefficient of $k \approx 0.6392$ N·m/rad. With a joint range of $\pm\pi/6$ radians, the corresponding torque span of a single spring unit is approximately ±0.3346 N·m. The torque response exhibits symmetrical hysteresis but without observable backlash. The system demonstrates good linearity, as evidenced by the low MSE and MAE values.

*3) Specifications & Parametric Design*

TABLE II
DESIGN PARAMETERS AND DELTA6 SPECIFICATIONS

| Design Parameters | | Specifications | |
|---|---|---|---|
| $a$ | 72 mm | Dimension (Neutral) | $166.2 \times 187.0 \times 233.6$ mm |
| $b$ | 21.24 mm | TCP Workspace | $73.2 \times 65.2 \times 53.1$ mm |
| $c$ | 30 mm | Total Weight | 308 g |
| $l_a$ | 40 mm | 3-DOF Unit Weight | 75 g |
| $l_b$ | 120 mm | $F_x$ Range | $-14.41$ – 15.03 N |
| $\theta_{\text{offset}}$ | $\pi/6$ rad | $F_y$ Range | $-15.03$ – 15.03 N |
| $k$ | 0.64 N·m/rad | $F_z$ Range | $-25.02$ – 22.09 N |
| | | $M_{x,y,z}$ Range | ±0.38, 0.33, 0.43 N·m |
| | | Force Resolution | 0.0118 N |
| | | Torque Resolution | 0.0003 N·m |

We define the core design parameters of Delta6 as: $(a, b, l_a, l_b, \theta_{offset}, k)$. Table II presents the design parameters used in the current prototype, along with the computed specifications in the right-hand section.

Joint-space configurations were sampled over the interval $\theta_{e1} \in [-\pi/6, \pi/6]$ $(i = 1 \dots 6)$. Each sample was mapped to Cartesian space via the Delta6 forward-kinematics solver to delineate the reachable workspace reported in the table. The allowable wrench bounds were then obtained from formula (1-13). Force resolution was estimated concurrently by adding the encoder's minimum angular resolution ($3.835 \times 10^{-4}$ rad) to every joint angle and recording the worst-case increment. Because the mechanism is not symmetric about the $y$-axis, the attainable $F_x$ range is correspondingly asymmetric. Besides, due to the transformation specified in (12), the three torque limits are scaled differently.

TABLE III
DESIGN VARIANTS AND RESULTING PERFORMANCE

| ID | Design Variables | | | | | | Computed Specification | | | |
|---|---|---|---|---|---|---|---|---|---|---|
| | $a$ (mm) | $b$ (mm) | $l_a$ (mm) | $l_b$ (mm) | $\theta_{\text{off}}$ (rad) | $k$ (N·m rad$^{-1}$) | $\lvert F_{\max}\rvert$ (N) | $\lvert M_{\max}\rvert$ (N·m) | $F_{\text{res}}$ (N) | $M_{\text{res}}$ (N·m) |
| 0 | 72.0 | 21.24 | 40 | 120 | 0.524 | 0.639 | 25.10 | 0.580 | 0.011 87 | 0.000 31 |
| 1 | 36.0 | 10.62 | 20 | 60 | 0.524 | 0.639 | 50.21 | 0.580 | 0.023 75 | 0.000 31 |
| 2 | 72.0 | 21.24 | 40 | 120 | 0.524 | 0.959 | 37.65 | 0.870 | 0.017 81 | 0.000 46 |
| 3 | 36.0 | 10.62 | 20 | 60 | 0.524 | 0.959 | 75.31 | 0.870 | 0.035 62 | 0.000 46 |

**Note:** $\lvert F_{\max}\rvert$—max force norm; $\lvert M_{\max}\rvert$—max torque norm; $F_{\text{res}}$, $M_{\text{res}}$—worst-case resolutions.

To examine how scale and compliance affect Delta6's load range and resolution, we analyzed 4 design variants (Table III):

*ID 0 - Baseline*: Current prototype serves as the reference point.

*ID 1 - Half-scale geometry*: All linear dimensions are uniformly reduced to 50% of the baseline while the spring stiffness is unchanged, isolating pure size effects.

*ID 2 - Stiffer springs*: Geometry is identical to the baseline, but the spring coefficient $k$ is increased to 150 %, highlighting the influence of higher compliance stiffness.

*ID 3 - Half-scale with stiffer springs*: Combines the 50 % geometric down-scaling with the 150 % spring-stiffness increase, allowing assessment of interaction effects between size reduction and increased stiffness.

According to the results from Table III, the overall size has significant impact on measurement range: halving every geometric dimension (ID 1) doubles the peak force but halves the force resolution. Increasing spring stiffness $k$ scales load capacity and resolution proportionally (ID 2). Hence, a compact design with stiff springs (ID 3) suits applications that demand high load capacity, whereas a larger structure with softer springs is preferable when higher compliance and sensitivity are prioritized.

### *B. Model Training and Evaluation*

#### *1) Training*

To implement the end-to-end machine learning models, including Transformer, LSTM, and GRU, we adopted the experimental setup depicted in Fig. 6-a. An FR3 manipulator was commanded to follow a random trajectory spanning most of Delta6's workspace, while an OptoForce six-axis F/T transducer synchronously logged the wrench $\boldsymbol{W}$ in frame $\{E\}$. Signals were acquired at 50 Hz for 2,000 s, resulting in $10^5$ annotated samples $(t, \boldsymbol{\theta}_e, \boldsymbol{W})$. The dataset was partitioned in an 80:20 split for training and held-out testing. Hyperparameters were chosen by 3-fold rolling-origin time-series cross-validation with early stopping. For each candidate we trained with Adam (learning rate = 0.001), and MSE loss (batch size 32) for up to 30 epochs, and stopped early if the validation loss (MSE) did not improve for 5 epochs. The best setting was selected by the lowest average validation loss across folds and then retrained with a 10% hold-out for early stopping (patience = 10, up to 150 epochs), keeping the best checkpoint. All networks were implemented in PyTorch and trained on a single NVIDIA RTX 2080 Ti GPU. Owing to their compact architecture, each model converged within minutes; their final hyperparameters and wall-clock training times are summarized in Table IV.

TABLE IV
HYPERPARAMETERS, MODEL SIZE & COMPUTE COST

| Model | $d_{model}/d_h$ | $n_h$ | $N_L$ | $p_{drop}$ | $T_w$ | $n_{in}/n_{out}$ | #Params (k) | MACs (k) | Train Time (s) |
|---|---|---|---|---|---|---|---|---|---|
| Transformer | 64 | 4 | 2 | 0.2 | 25 | 6/6 | 102.4 | 2627.6 | 196.9 |
| LSTM | 128 | - | 2 | 0.3 | 25 | 6/6 | 202.5 | 4992.8 | 229.1 |
| GRU | 128 | - | 2 | 0.2 | 25 | 6/6 | 152.1 | 3744.8 | 162.5 |

*Note:* MACs are estimated by counting dominant matrix multiplications per forward pass.

#### *2) Evaluation*

- *Inference Time & Latency*

Fig. 7 compares the inference latency of all four approaches on four representative hardware platforms: a desktop PC (Intel Core i7-12700K + RTX 2080 Ti), two edge devices (Raspberry Pi 5 and Jetson Nano Super), and an MCU (Arduino Nano Every). On the MCU, only the analytical solution can be deployed, yet it still completes a full inference in ≈ 10 ms; when the additional overhead of encoder data acquisition and transfer is included, the end-to-end loop maintains a stable rate of ≥ 50 Hz. Among the edge devices, the Raspberry Pi 5 CPU keeps any of the three learned models below 100 ms, corresponding to a reliable 10 Hz update rate, whereas the Jetson Nano Super leverages its CUDA cores to sustain ≥ 50 Hz for all models. On the desktop platform, the three neural-network models all converge to solutions in roughly 1 ms.

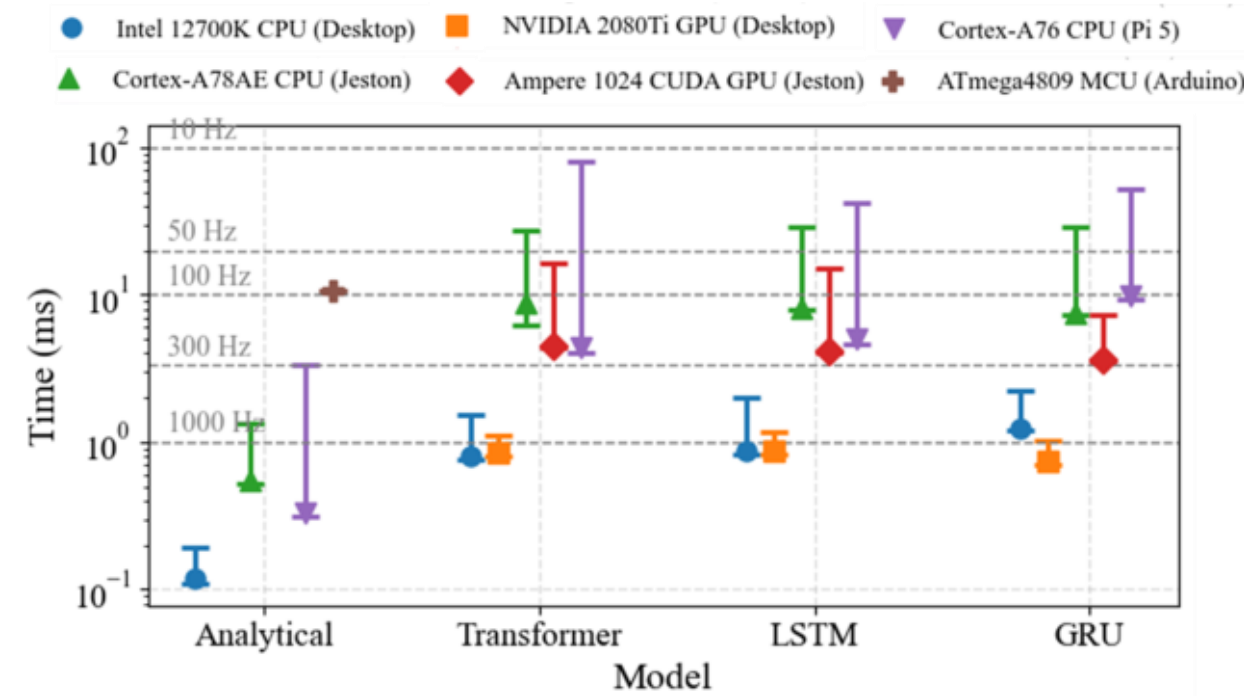

**Fig. 7.** Inference Time Comparison of Different Methods on CPU and GPU

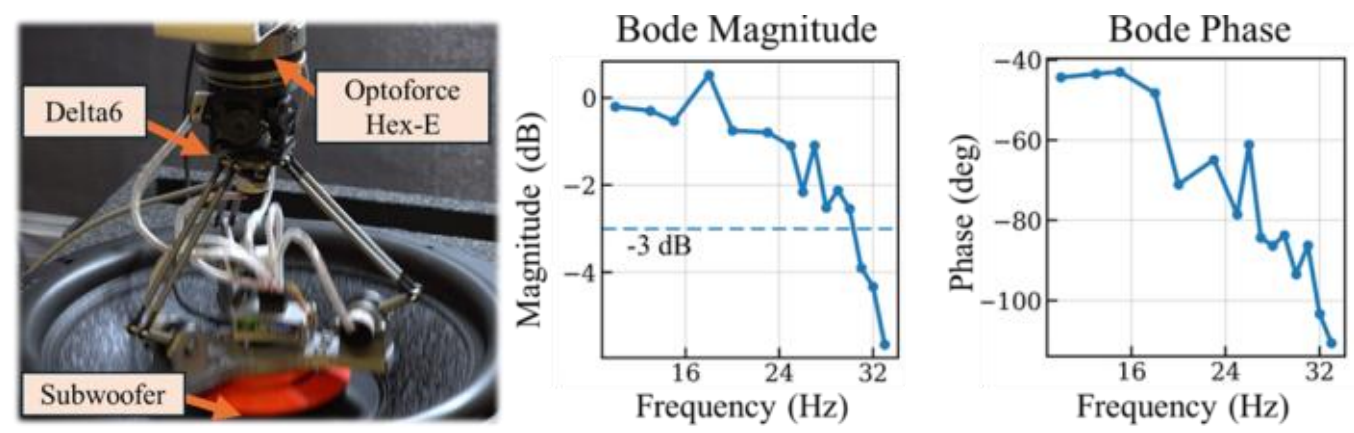

**Fig. 8.** Delta6's Bode Magnitude and Phase Relative to OptoForce.

In deployment, communication and scheduling dominate. For the desktop-based prototype using a 115200-baud CDC-ACM link with a 6-byte request and a 16-byte reply, the wire time is about 2 ms and the typical round trip with USB scheduling is 3–5 ms. With a 6.6 ms sensor scan, 6.6 ms host polling, and about 1 ms inference on CUDA, the end-to-end latency ranges for using Transformer model from about 10.6 ms in the best case with event-aligned acquisition and computation to about 25.8 ms in the worst case. Lower latency can be achieved by upgrading the MCU, replacing round-robin polling with synchronized sensor queries, and increasing the serial baud rate.

- *Dynamic Response*

To characterize the relative dynamic response of Delta6, we used the setup shown on Fig. 8. Delta6 was inverted and rigidly mounted on a high-power subwoofer that produced approximately sinusoidal motion from 10 to 100 Hz with about 4 mm peak-to-peak displacement. The force-sensing end of Delta6 was coupled to an OptoForce sensor, which served as the reference for comparison. We applied single-frequency sine excitations at each test frequency and recorded the $F_z$ waveforms. OptoForce was sampled at 333 Hz and Delta6 at 150 Hz. Applying excitation responses at frequencies between 10 and 33 Hz, four methods were tested and showed similar results. The Bode plots of analytical estimator are shown on Fig.8, with -3dB cutoff occurring near 30 Hz and a phase lag of about −90°.

- *Per-Axis Accuracy*

An evaluation dataset was collected using the same procedure as the training dataset. For each model, absolute percentage error of F/T relative to the sensor's measurement range for each axis is summarized in Table V. The Transformer achieves the lowest error on 17 of the 18 axis–percentile combinations. The sole exception is $M_z$ at the untrimmed maximum, where the Analytical baseline is lower, likely because the Analytical method's spring-stiffness parameter was calibrated by data from z-axis. At the 99th percentile, the Transformer bounds the worst-case error below 3.8% FS on every axis, whereas the Analytical model remains below 7.1%. Overall, the Transformer not only reduces typical errors but also tightens the error tail.

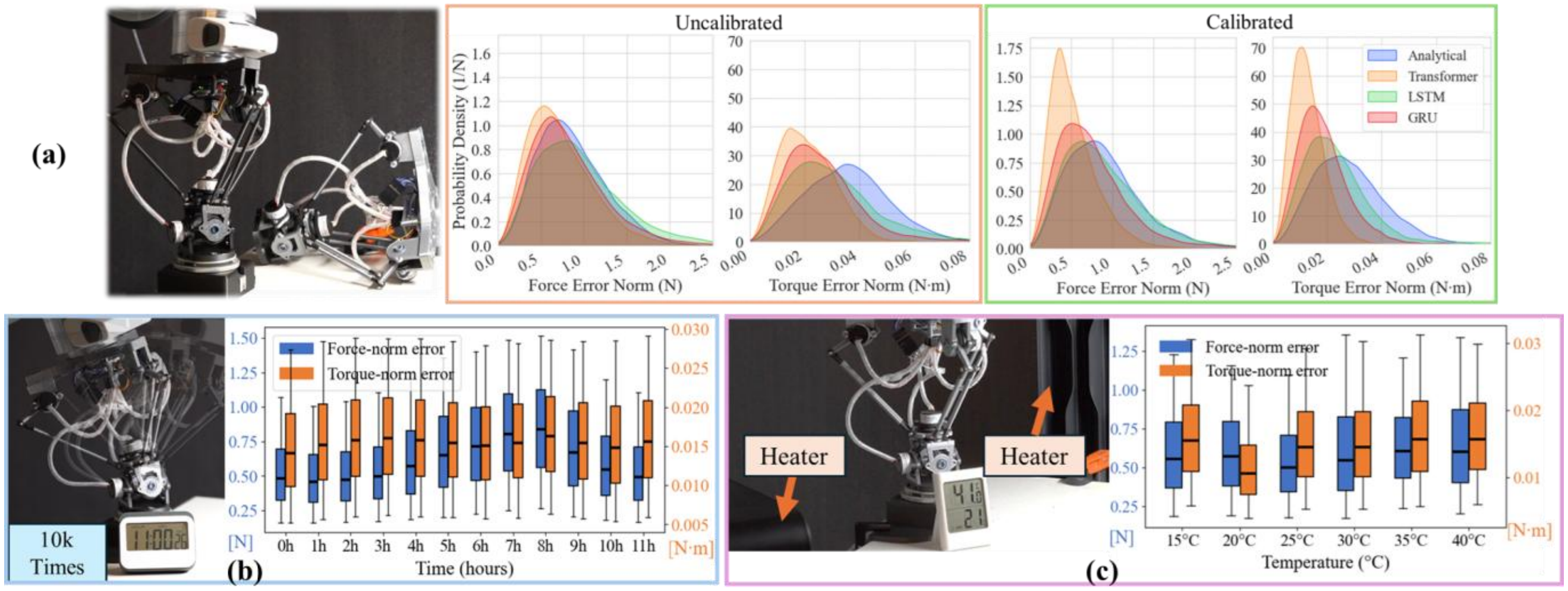

**Fig. 10.** Robustness Evaluation of the Force/Torque Estimator. a) Experimental Setup and Force/Torque Error Distributions on Calibrated vs. Uncalibrated Hardware. b) Durability Test: Error Drift Across 10,000 Cycles. c) Thermal Robustness: Error Distribution at 15 – 40 °C.

TABLE V
MAXIMUM FORCE/TORQUE ERROR PERCENTAGES

| | | Force | | | Torque | | |
|---|---|---|---|---|---|---|---|
| **Model** | **Percentile** | **Fx (%)** | **Fy (%)** | **Fz (%)** | **Mx (%)** | **My (%)** | **Mz (%)** |
| Analytical | 100th | 7.16 | 9.76 | 6.39 | 11.12 | 9.44 | **5.67** |
| | 99th | 5.06 | 6.50 | 3.64 | 7.08 | 6.92 | 4.34 |
| | 95th | 3.81 | 4.62 | 2.58 | 5.45 | 5.41 | 3.42 |
| Transformer | 100th | **4.41** | **6.30** | **4.06** | **6.39** | **4.81** | 8.63 |
| | 99th | **2.72** | **2.10** | **2.65** | **3.78** | **2.68** | **2.21** |
| | 95th | **1.78** | **1.45** | **1.99** | **2.90** | **2.03** | **1.55** |
| LSTM | 100th | 8.95 | 8.78 | 7.65 | 9.66 | 11.40 | 11.77 |
| | 99th | 5.73 | 5.22 | 4.48 | 4.98 | 5.08 | 7.68 |
| | 95th | 3.91 | 3.44 | 3.11 | 3.66 | 3.46 | 4.89 |
| GRU | 100th | 6.66 | 8.19 | 7.23 | 7.84 | 7.96 | 9.47 |
| | 99th | 3.99 | 4.76 | 4.15 | 4.54 | 3.72 | 5.40 |
| | 95th | 2.89 | 2.86 | 2.77 | 3.39 | 2.70 | 3.70 |

- *Cross-Axis Sensitivity*

Using the setup in Fig. 6, cyclic single-axis excitations were applied while other axes were held quiescent. For each excited axis, we constructed an FS-normalized residual sensitivity matrix by regressing the Delta6-OptoForce difference against the on-axis OptoForce signal, both normalized to full scale. Diagonals report residual on-axis gain error, while off-diagonals report cross-axis sensitivity, both of which are ideally zero. Fig. 9 presents the heatmaps for the Analytical and Transformer estimators. Quantitatively, the mean off-diagonal decreases from 0.048 FS/FS for the Analytical model to 0.019 FS/FS for the Transformer, indicating a markedly lower cross-axis sensitivity under the residual metric. On the same dataset, we further analyzed repeatability for the two estimators. Across six axes, the median per-cycle gain standard deviation decreased from 0.0049 to 0.0032 (FS-normalized, FS/FS).

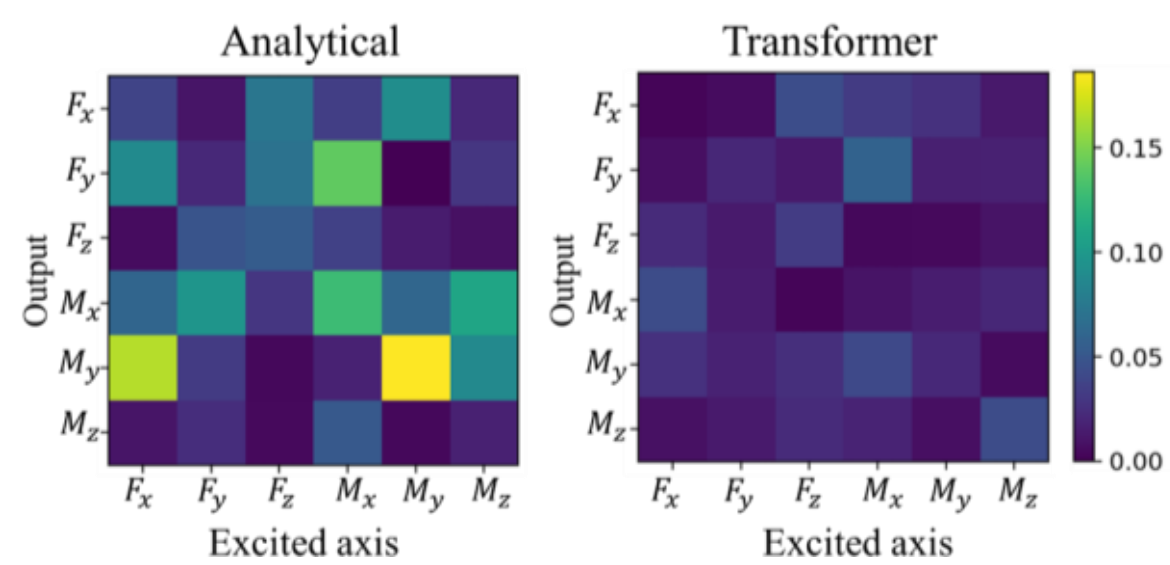

**Fig. 9.** FS-normalized Residual Cross-axis Sensitivity Heatmaps.

Post-test inspection indicates compliance and backlash cause model–plant mismatch. The Transformer evidently learns these effects, which are hard to encode analytically.

*3) Robustness Tests*

- *Zero-calibration Validation*

To evaluate the generalizability of the learned models to uncalibrated hardware. We assembled a second, uncalibrated Delta6 prototype (black). Fig.10-a contrasts the F/T error–norm densities for the uncalibrated unit with the calibrated reference. For the uncalibrated prototype, the learning-based models still achieve lower error densities than the analytical formulation, although the gap is narrower. This suggests that the networks have implicitly captured error sources beyond assembly tolerances, such as hysteresis in the compliant elements and friction or backlash. For the calibrated Delta6, the advantage is clearer: every learned method surpasses the analytical solution, with the Transformer exhibiting the most concentrated error distribution.

- *Durability & Thermal*

Delta6 was subjected to a long-duration durability test in which a new sampled Cartesian command was issued every 4 s for 10,000 consecutive cycles ($\approx$ 11 h). Sensor data were logged at 50 Hz, yielding $2 \times 10^6$ samples and trajectory was partitioned into eleven one-hour segments. Force- and torque-norm errors were obtained by comparison with the OptoForce sensor, and their distributions are summarized in Fig. 10-b. Using the Transformer model for inference, we observed no monotonic growth in error throughout the experiment: although a transient increase occurred between hours 4 and 8, likely caused by joint break-in, the error later returned to its initial level. Over the entire 10,000-cycle test, the drift rate remained negligible: 3.240% FS/h for force norm and 0.002% FS/h for torque norm, indicating that the prototype's accuracy was not degraded.

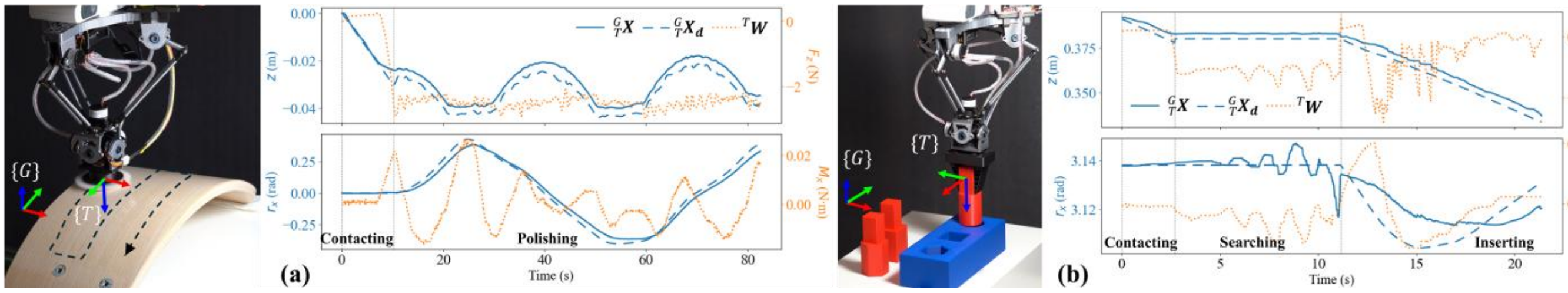

**Fig. 11.** Contact-rich task demonstrations using Delta 6. (a) Force-controlled wood buffing on a curved panel; (b) force-guided tight-clearance insertion.

Thermal robustness was evaluated by stepping the ambient temperature from 15 °C to 40 °C using two space heaters. After thermal equilibrium was achieved at each set-point, 600 s of data were collected and processed identically to the durability study. The error-norm box plots (Fig. 10-c) reveal no discernible positive correlation between temperature and error, demonstrating that the estimator sustains uniform accuracy across the tested thermal range typical of laboratory conditions: the drift slope was 0.007% FS/°C for force norm and 0.012% FS/°C for torque norm.

### *C. Applications Validation*

The impedance controller was configured with zero virtual mass $\boldsymbol{M} = \boldsymbol{0}_{6\times 6}$ , $\boldsymbol{B} = diag(25, 25, 25, 0.8, 0.8, 0.8)$ , and $\boldsymbol{K} = diag(500, 500, 1000, 10, 10, 10)$. All values are reported in the customary units $m$, $rad$, $s$, and $N$.

#### *1) Wood Buffing on Unmodelled Curved Surface*

To evaluate practical performance, the Delta6 was equipped with a rotary buffing pad and commanded to follow the curved wooden panel shown in Fig. 11-a. The experimental protocol followed the procedure described in Sec. IV-B. The polishing spindle was first run about 8,000 rpm. Once the TCP contacted the workpiece, the Cartesian-impedance controller switched to polishing mode and imposed a constant normal-force set-point of $F_{z,ref}$ as -2.5N, in frame $\{T\}$. A zig-zag toolpath was generated by the trajectory planner to cover the surface. The task was completed in 85s without loss of contact. The $z - F_z$ traces verify that the controller maintained the prescribed normal force throughout, while the $r_x - M_x$ traces indicate that unwanted torques were kept near zero, ensuring the tool always remained tangential to the curved surface.

#### *2) Peg-in-Hole Tight Assembly*

To quantify precision-assembly capability, the Delta 6 tool was replaced by a gripper and tasked with inserting pegs of three geometries (cylindrical, square, and hexagonal) into matching holes, as illustrated in Fig. 11-b. Hole clearances were less than 0.2 mm, rendering pure position control ineffective. The controller therefore executed the three-stage routine of Sec. IV-C: after initial contact, a compliant search phase aligned the peg, followed by force-guided insertion. For the representative cylindrical case, the $z/F_z$ traces reveal first contact at $t \approx 3s$, an eight-second search culminating in a sharp increase in $F_z$, that marks hole acquisition, and a progressive force decline as the peg seats fully. Simultaneously, the $r_x/M_x$ traces show unwanted torques actively suppressed, allowing continuous micro-re-orientation and keeping the gripper coaxial with the bore. Across ten trials per geometry, the force-guided strategy achieved insertion rates of 10/10 for the cylinder, 8/10 for the square peg, and 7/10 for the hexagon. With the force sensor disabled, success plummeted to 1/10, 0/10, and 0/10, respectively. These results underscore the necessity of real-time wrench sensing tight clearance assembly and demonstrate the controller's robustness across varied contact geometries.

## VI. CONCLUSION

In conclusion, Delta6 is a modular, compliant, magnetic-encoder end-effector that delivers accurate, robust, and cost-effective 6-DoF wrench sensing. Future work will focus on (i) simplifying assembly by adopting snap-fit or self-indexing joints that eliminate most fasteners, (ii) enhancing durability and stiffness by replacing high-wear printed parts with aluminum or carbon-fiber/composite components, and (iii) achieving fully onboard force–torque estimation by porting a compact Transformer-based inference model to MCU-class processors.

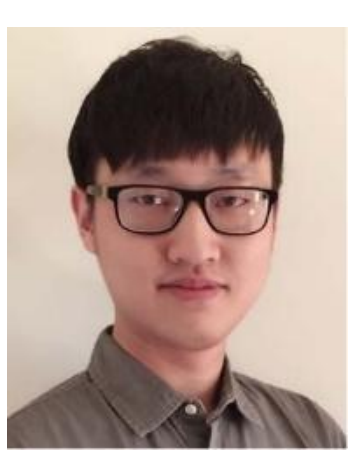

**Yue Feng** (Member, IEEE) received the B.S. degree in Mechanical Engineering from the University of Delaware, Newark, DE, USA, in 2018, and the M.S. degree from Nanyang Technological University (NTU), Singapore, in 2019. From 2019 to 2022, he was with the Robotics Research Center, NTU, as a Research Associate. He is currently pursuing a Ph.D. degree in robotics at NTU. His research interests include mechatronics, robot teleoperation, and robot learning.

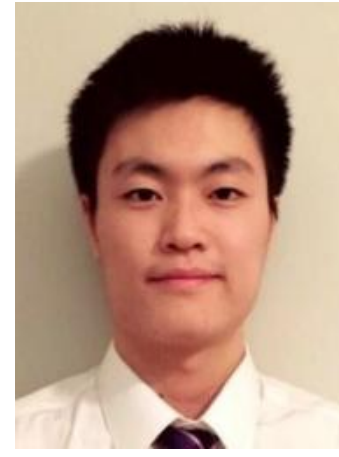

**Weicheng Huang** received his B.S. degree in Computer Engineering from Ohio State University. He is currently pursuing an M.S. degree in Computer Science at Northeastern University. From 2016 to 2023, he worked as a Senior Software Engineer at Amazon and Meta, focusing on large-scale distributed systems and software architecture. His research interests include machine learning, cloud computing, and high-performance computing systems.

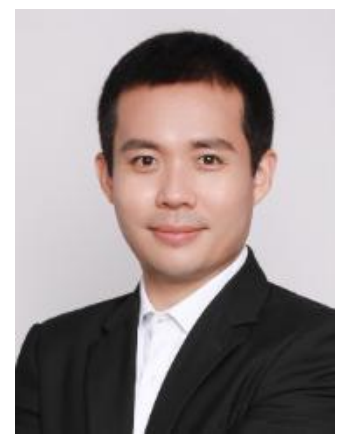

**Huixu Dong** received the B.Sc degree in mechatronics engineering from Harbin Institute of Technology in China, in 2013 and obtained Ph.D. degree at Robotics Research Centre of Nanyang Technological University, Singapore 2018.
He was a post-doctoral fellow in Robotics Institute of Carnegie Mellon University and National University of Singapore. He is an associate editor of IEEE Robotics and Automation Letter (IEEE RA-L), IEEE Transactions on Automation Science and Engineering (IEEE T-ASE), ICRA 2023-2025, IROS 2022-2025 and AIM 2022-2024. His current research interests include robotic perception and manipulation in unstructured environments, robotic gripper/hand.

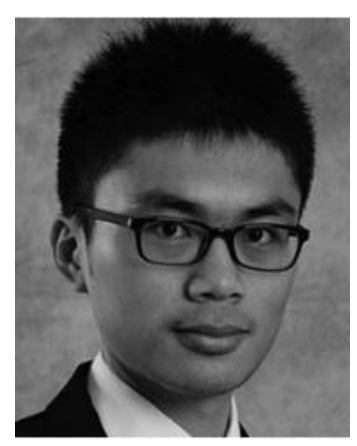

**Chen Qiu** received his B.S. degree in spacecraft design and engineering from Beihang University in Beijing, China, in 2011. He obtained his Ph.D. in robotics from the Centre for Robotics Research at King's College London in the United Kingdom in 2016. Then, he held the position of Research Fellow at the Robotics Research Center of Nanyang Technological University in Singapore. His research interests encompass the kinematics and dynamics of robotics, compliant mechanisms and robotics, and medical and humanoid robots.

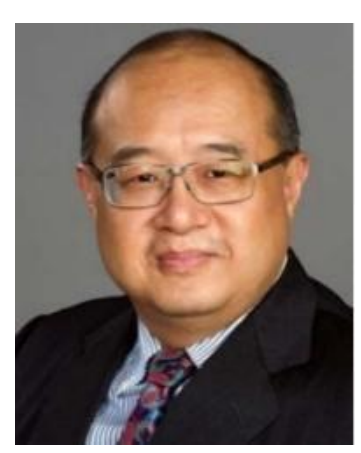

**I-Ming Chen** (S'90–M'95–SM'06-F'14) received B.S. degree from National Taiwan University, Taipei, Taiwan, in 1986, and M.S. and Ph.D. degrees from the California Institute of Technology, Pasadena, in 1989 and 1994, respectively. He has been with the School of Mechanical and Aerospace Engineering of Nanyang Technological University (NTU) in Singapore since 1995. He is Currently Co-Director of CARTIN (Centre for Advanced Robotics Technology Innovation), and Senior Editor of IEEE Transactions on Robotics and Automation Practice. His research interests are in construction and logistics robots, wearable devices, human-robot interaction and industrial automation. Professor Chen was Editor-in-chief of IEEE/ASME Transactions on Mechatronics from 2020 to 2022, and General Chairman of 2017 IEEE International Conference on Robotics and Automation (ICRA 2017) in Singapore. He is also Fellow of ASME, and Fellow of Singapore Academy of Engineering (SAEng).